\documentclass[lettersize,journal]{IEEEtran}
\usepackage{amsmath,amsfonts}
\usepackage{algorithmic}
\usepackage{algorithm}
\usepackage{array}
\usepackage[caption=false,font=footnotesize,labelfont=rm,textfont=rm]{subfig}
\usepackage{textcomp}
\usepackage{stfloats}
\usepackage{url}
\usepackage{verbatim}
\usepackage{graphicx}
\usepackage{cite}
\usepackage{graphicx}
\usepackage{amsmath}
\usepackage{amssymb}
\usepackage{booktabs}
\usepackage{multirow}
\usepackage{utfsym}
\usepackage{makecell}
\usepackage{foreign}
\usepackage[pagebackref,breaklinks,colorlinks]{hyperref}
\usepackage[capitalize]{cleveref}
\begin{document}

\title{Video-Specific Query-Key Attention Modeling for \protect\\ Weakly-Supervised Temporal Action Localization}


\author{Xijun Wang$^1$, and Aggelos K. Katsaggelos$^2$\\
$^1$Dept. of Computer Science, Northwestern University, Evanston, IL, USA\\
$^2$Dept. of Electrical and Computer Engineering, Northwestern University, Evanston, IL, USA\\
}



\maketitle

\begin{abstract}
Weakly-supervised temporal action localization aims to identify and localize the action instances in the untrimmed videos with only video-level action labels. When humans watch videos, we can adapt our abstract-level knowledge about actions in different video scenarios and detect whether some actions are occurring. In this paper, we mimic how humans do and bring a new perspective for locating and identifying multiple actions in a video. We propose a network named VQK-Net with a video-specific query-key attention modeling that learns a unique query for each action category of each input video. The learned queries not only contain the actions' knowledge features at the abstract level but also have the ability to fit this knowledge into the target video scenario, and they will be used to detect the presence of the corresponding action along the temporal dimension. To better learn these action category queries, we exploit not only the features of the current input video but also the correlation between different videos through a novel video-specific action category query learner worked with a query similarity loss. Finally, we conduct extensive experiments on three commonly used datasets (THUMOS14, ActivityNet1.2, and ActivityNet1.3) and achieve state-of-the-art performance. 
\end{abstract}

\let\thefootnote\relax\footnote{This work has been submitted for possible publication. Copyright may be transferred without notice, after which this version may no longer be accessible.}


\section{Introduction}
\label{sec:intro}
In recent years, video analysis has been a rapidly developing topic due to the explosive growth of video data used in various real-world applications, especially in the field of video temporal action localization (TAL). The reason for this is that long untrimmed videos contain more interesting foreground activity and useless background activity, and they are more common than short trimmed videos. TAL is a highly challenging task that aims at predicting the start and end times of all action instances and identifying their categories in untrimmed videos. Many works have been done in a fully-supervised manner, where both the video-level labels and the temporal boundary annotations are provided during training \cite{Long_2019_CVPR, xu2020g, Xia_2022_CVPR, zhu2022learning}. In contrast, the weakly-supervised temporal action localization (WTAL) task attempts to rely only on video-level labels to localize action instances, which can significantly relieve the high cost of manually annotating the temporal boundaries. 

\begin{figure}[t]
\centering
\includegraphics[width=1.0\linewidth]{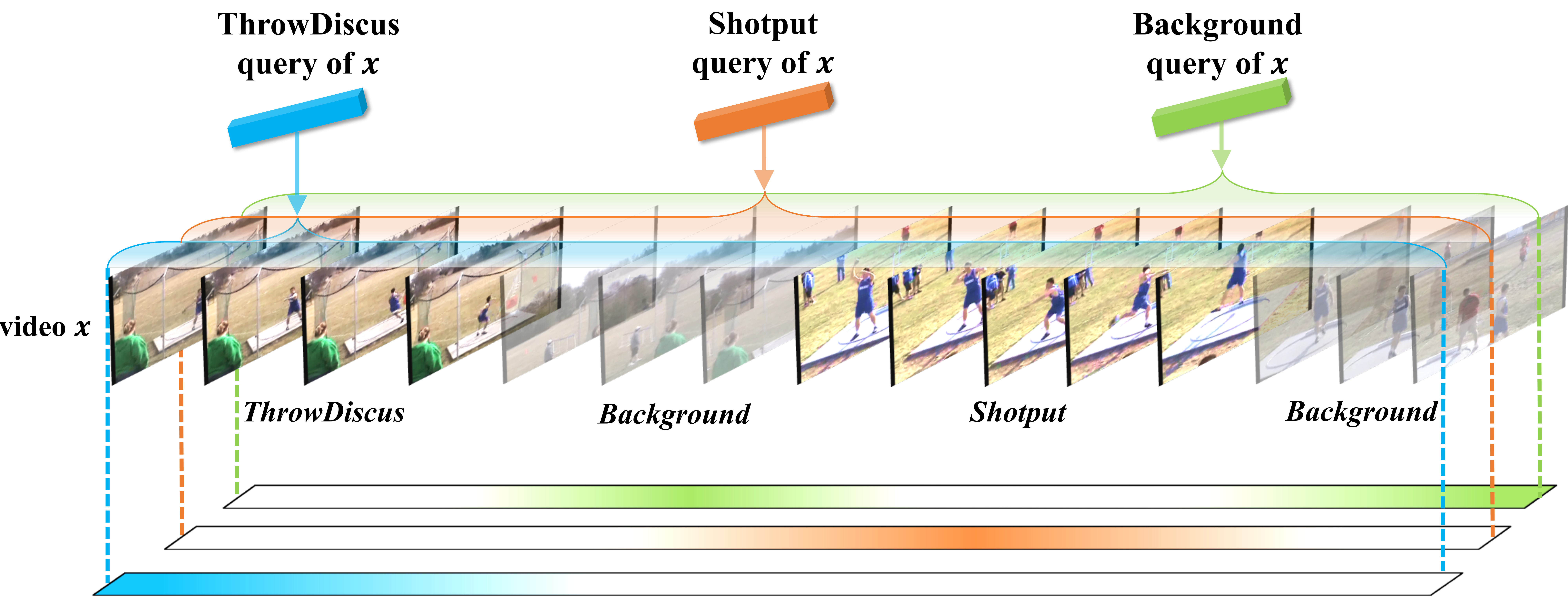}
\caption{Action category queries are learned so as to contain action knowledge at the abstract level, which can be used to identify and detect corresponding actions in the target video.}
\label{fig1}
\label{fig:idea}
\end{figure}

In common with other weakly-supervised video understanding tasks \cite{feng2021mist,hong2020mini,sultani2018real}, many WTAL methods adopt a multiple instance learning (MIL) strategy \cite{paul2018w,rashid2020action,lee2020background,hong2021cross,islam2021hybrid,li2022exploring}. With this strategy, one first computes segment-level class probability scores, then aggregates the top scores for each class as the video-level class scores, and then forms the video-level classification losses to perform the optimization with the given video-level labels. 

With the success of Transformer \cite{vaswani2017attention} in many computer vision tasks \cite{arnab2021vivit,dosovitskiy2020image}, some recent TAL works build their models based on Transformer’s encoder-decoder framework \cite{nawhal2021activity,liu2021end,li2022w} and achieve good results. However, all these works aim to learn a set of action queries corresponding to the latent representations of a set of time areas (action proposals). Few works attempt to solve the TAL task in such a way that the abstract-level knowledge of each action category is learned and used to recognize and detect the corresponding actions in various video scenes, just like how humans do. The closest work to this idea is STPN\cite{nguyen2018weakly}. However, it is limited to learning a uniform set of weight parameters for action categories using a fully connected layer. 

In this work, we present a new video-specific query-key attention mechanism and propose our VQK-Net model based on it. Our high-level idea is illustrated in \cref{fig:idea}. More specifically, we propose to learn the video-specific action category queries that can be adapted in different video scenarios and simultaneously maintain the action core knowledge features used to detect actions in the videos, i.e., the learned action category queries contain abstract knowledge of actions, and they are tailored to the target video scene to optimize the application of this knowledge. To accomplish this, we propose incorporating video features into the action category queries learning process for two reasons: 1) Since the same action can appear differently in different videos, integrating input video information could help learn the action category queries to better fit into different video scenarios. 2) Some action knowledge can be hidden in the video. Therefore, video features can help the model learn the action's core knowledge features. We achieve this by referring to the cross-attention in Transformer's decoder design.

However, so far, we have overlooked one problem, video features can confuse the model in learning the action category queries whose corresponding actions do not occur in the input video, i.e., there is no useful information about these categories in the input video. Therefore, we proposed a query similarity loss to tackle this problem. Our idea is that for any two videos containing the same action category, their corresponding action category queries should look similar because they both learn the abstract-level knowledge features of that action. With query similarity loss, on the one hand,  we can compel the model to learn the action's core knowledge features by leveraging correlations between videos of similar action categories. On the other hand, in addition to using the video-level classification loss to explicitly teach the model what action categories exist in the video, the query-similarity loss implicitly provides the model with similar information. Since for the actions that occur in the video, the corresponding action category queries will be better learned and enhanced under the guidance of query similarity loss, which in turn suppresses the effect of action category queries that are not present in the video.

A two-stage model training strategy is commonly adopted in solving the TAL task. In the first feature extraction stage, a pre-trained feature extractor (\textit{e.g.}, I3D \cite{carreira2017quo}), which is typically trained on a large trimmed dataset (\textit{e.g.}, Kinetics) for the general video action classification tasks, is used to extract the video features from the untrimmed video input. In the second stage, a temporal localization model is then trained using the extracted video features. In this paper, we also follow this two-step training strategy.

To summarize, our main contributions are as follows: 
\begin{itemize}
    \item We propose the VQK-Net model with a video-specific query-key attention modeling, where the model learns a unique query for each action category of each input video and uses these learned action category queries to identify and detect the corresponding actions from the video. 
    \item We design a novel video-specific action category query learner worked with a query similarity loss. The learned queries contain the actions' abstract knowledge features to detect and identify the actions. To best apply the learned knowledge in different video scenarios, the queries are learned to adapt to the target video scene. 
    \item We conduct extensive experiments on our design, and our proposed method achieves state-of-the-art performance on benchmark datasets.
\end{itemize}

\section{Related Work}
Thanks to the powerful representation capabilities of deep learning and the existing large-scale video datasets\cite{caba2015activitynet, kay2017kinetics, monfort2019moments}, notable progress has been achieved in the video action recognition field. Many works adopt the two-stream network design \cite{simonyan2014two}, which incorporates the optical flow \cite{horn1981determining} as a second stream, in addition to the RGB stream.

As the demand for video analytics in modern life continues to grow, the research interest has expanded from trimmed video action classification to video TAL of more common untrimmed videos.
To relieve the expensive cost of acquiring precise time stamp annotations, researchers expanded their attention from TAL to WTAL which relies only on video-level labels. UntrimmendNet \cite{wang2017untrimmednets} proposes to predict the segment-level classification score, STPN \cite{nguyen2018weakly} further introduces a sparsity loss and class-specific proposals. AutoLoc\cite{shou2018autoloc} introduces the outer-inner contrastive loss to find the temporal boundaries effectively. W-TALC \cite{paul2018w} develops a multiple instance learning scheme and has been used in many works. Among them, some works \cite{islam2020weakly,paul2018w} directly calculate the video-level class score by aggregating the predicted segment-level class scores, in which the background activity is not explicitly considered and can be misclassified as foreground activities. To address this issue, HAM-net \cite{islam2021hybrid} chooses to suppress the background segments and improve the results by learning the segment-level foreground probability distribution. DGAM \cite{shi2020weakly} used a conditional variational auto-encoder to separate the nearby context frames from the actual action frames,  UM\cite{lee2021weakly} utilized the magnitude difference between the foreground and background features, and DELU \cite{chen2022dual} targets reducing the action-background ambiguity by utilizing evidential deep learning. In addition, some WTAL works \cite{rashid2020action, yang2022acgnet} utilize graphs to model the relationship between video segments. W-ART \cite{li2022w} follows the transformer encoder-decoder approach \cite{vaswani2017attention} to learn action queries for the action proposals. Some work \cite{zhai2020two,huang2022weakly, he2022asm} utilize the pseudo labels to refine their networks, in which RSKP\cite{huang2022weakly} learns the snippets’ cluster centers from the given video datasets for pseudo-label generation.  In this paper, we propose to solve the WTAL problem by mimicking how humans detect and identify an action instance from a video. Our model learns the video-specific action queries, which contain abstract knowledge to detect and identify action instances from videos, while these queries can be adapted to different video scenarios.

\section{Proposed Method}
\label{sec:proposed_method}
In this section, we present a comprehensive explanation of the proposed VQK-Net model for WTAL. We first formulate the WTAL problem in \cref{sec:prob_state} and describe the feature extraction in \cref{sec:feat_extact}. Then we provide an overview of the main pipeline of VQK-Net in \cref{sec:main_pipe}. After that, we delve into the key components of the model: query learner and query similarity loss in \cref{sec:query_learner}, and query-key attention module in \cref{sec:query-key-att}. Finally, we detail the training objective functions in \cref{sec:train_objects} and how the temporal action localization is performed in \cref{sec:temp_act_loc}. The overview of our model is shown in \cref{fig:framework}.
\subsection{Problem Statement}
\label{sec:prob_state}
 We formulate the WTAL problem as follows: During training, for a video \(\textbf{x}\), only its video-level label is given, denoted as \(\textbf{y} = [y_1, y_2, ..., y_{C+1}]\),  where C+1 is the number of action categories and the \((C+1)\)-th class is the background category. An action can occur multiple times in the video, and \(y_i = 1\) only if there is at least one instance of the \(i\)-th action category in the video. During testing, given a video \(\textbf{x}\), we aim at detecting and classifying all action proposals temporally, denoted as \(\textbf{x}_{pro} = \{(t_s^j, t_e^j, c^j, \varepsilon^j)\}_{j=1}^{r(\textbf{x})}\), where \(r(\textbf{x})\) is the number of action proposals for video \(\textbf{x}\), and \(t_s^j, t_e^j, c^j, \varepsilon^j\) denote the start time, the end time, the predicted action category and the classification score of the predicted action category, respectively.

\subsection{Feature Extraction}
\label{sec:feat_extact}
Following the previous work in \cite{paul2018w}, for each input video \(\textbf{x}\), we split it into multi-frame segments, each segment containing a fixed number of frames. To handle the variation of video lengths, a fixed number of \(T\) segments are sampled from each video. Following the two-stream strategy used in action recognition \cite{carreira2017quo,feichtenhofer2016convolutional}, we extract the segment-level RGB and flow features vectors \(\textbf{x}_{rgb}\in\mathbb{R}^{D/2}\) and \(\textbf{x}_{f}\in\mathbb{R}^{D/2}\) from a pre-trained extractor,  i.e., I3D, with dimension \(D=2048\). At the end of the feature extraction procedure, each video \(\textbf{x}\) is represented by two matrices \(X_{rgb}\in\mathbb{R}^{{T\times (D/2)}}\) and \(X_{f}\in\mathbb{R}^{{T\times (D/2)}}\), denoting the RGB and flow features for the video, respectively.

\subsection{Main Pipeline Overview}
\label{sec:main_pipe}
\begin{figure*}[t]
\centering
\includegraphics[width=2.0\columnwidth]{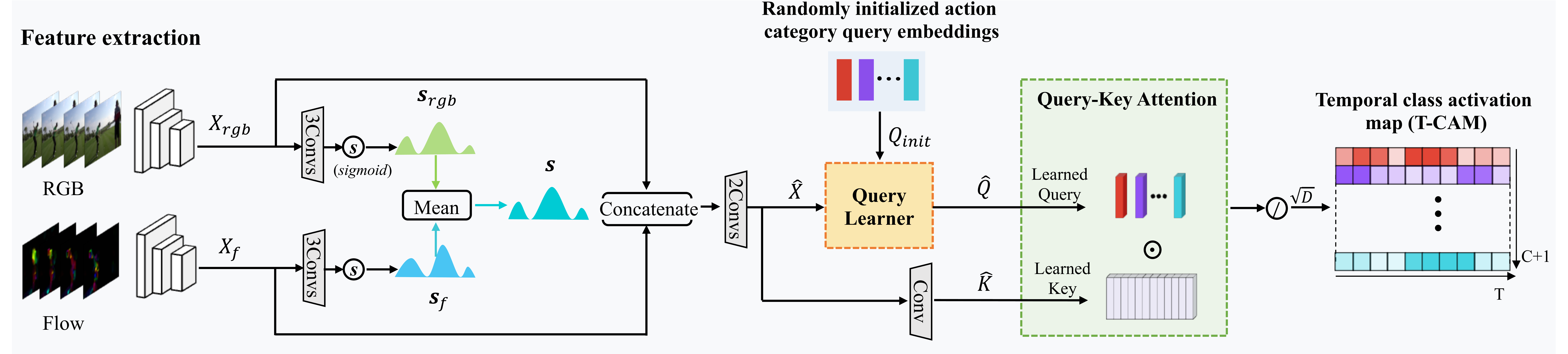}
\caption{Overview of our proposed VQK-Net model. $\otimes$, $\oslash$ and $\odot$ denote the element-wise multiplication, element-wise division and vector inner product.}
\label{fig:framework}
\end{figure*}

\cref{fig:framework} shows the main pipeline of our proposed VQK-Net model. For an input video \(\textbf{x}\), we refer to the mutual learning scheme \cite{hong2021cross} to learn the probability of each segment being foreground from two stream features \({X}_{rgb}\) and \({X}_f\): as shown in \cref{fig:framework}, we first employ three convolution layers with LeakyRelu activations in between and a sigmoid function on \({X}_{rgb}\) to get the segment-level foreground probability distribution \(\textbf{s}_{rgb}\in\mathbb{R}^T\), and the same to obtain \(\textbf{s}_f\in\mathbb{R}^T\) with \({X}_{f}\) .  
We average them to get the final \(\textbf{s}\in\mathbb{R}^T\): \(\textbf{s} =  \frac{\textbf{s}_{rgb} + \textbf{s}_f}{2}\). 

Then, we first directly concatenate RGB and flow features in the feature dimension, i.e., concatenate \({X}_{rgb}\) and \({X}_f\) to form \(X\in\mathbb{R}^{T\times D}\), and input \(X\) to two convolution layers with LeakyReLU activations in between to learn the final fusion feature \(\hat{X}\in\mathbb{R}^{T\times D}\).
The query learner module then takes \(\hat{X}\) and C+1 randomly initialized learnable action category query embeddings, which can be stacked to form a category query matrix \(Q_{init}\in\mathbb{R}^{(C+1)\times D}\),  as inputs. In this module, we refer to the Transformer decoder's design \cite{vaswani2017attention} with our proposed query similarity loss to learn the final category query matrix \(\hat{Q}\in\mathbb{R}^{(C+1)\times D}\), which contains the learned action category queries for C+1 classes. Finally, we feed \(\hat{X}\) through a convolution layer to learn the final video features \(\hat{K}\in\mathbb{R}^{T\times D}\) of the input video, used as the video key. The learned query matrix \(\hat{Q}\) and learned key matrix \(\hat{K}\) will be input to the following query-key attention module to produce the temporal class activation map (T-CAM) \(A\in\mathbb{R}^{(C+1)\times T}\). The details are discussed in the following Sections.

\subsection{Query Learner}
\label{sec:query_learner}
The query Learner is an essential part of our VQK-Net model. It learns the video-specific action category queries by exploiting both the video features and the correlation between different videos. The final learned queries will be used to query and detect the corresponding actions along the temporal dimension in the input video.\\

\noindent
\textbf{Structure.}    As we explain in \cref{sec:intro}, different videos have different scenarios, so it is beneficial to learn the video-specific action category queries that can best match the input video. Given the input video \(\textbf{x}\),  we proposed to include the input video features \(\hat{X}\) into learning action category queries instead of just learning the action category queries for all the videos based on \(Q_{init}\). In addition, learning action category queries for specific videos provides the possibility of using correlations between videos to further enhance the learned action category queries.

To include the features learned from the input video, we refer to the Transformer decoder's design. The head attention operation function \(f_{h}(\cdot)\) used in our query learner is defined as:
\begin{equation}
\label{eq:headattn}
f_{h} (Q, K, V) = HW_O,
\end{equation}
where
\begin{equation}
\label{head}
H = f_a(QW_Q, KW_K, VW_V),
\end{equation}and
\begin{equation}
\label{eq:attn}
f_a (Q, K, V) = \varsigma(\frac{QK^\top}{\sqrt{D}})V.
\end{equation}
\(Q, K, V\) are three input matrices, and \(W_Q, W_K, W_V\) and \(W_O\in \mathbb{R} ^{D\times D}\) are learnable parameter matrices. \(\varsigma(\cdot)\) takes a matrix as input, and it denotes that each row of its input is normalized using the softmax operation.

As shown in \cref{fig:q-k}(a), in our query learner, a head attention operation will first operate on the initial action category query matrix \(Q_{init}\)  itself, i.e., \(f_{h}(Q_{init}, Q_{init}, Q_{init})\). After that, a residual connection and Layer Normalization will be used to output \(Q_{1}\). The video feature \(\hat{X}\) will be used in the second head attention operation to adapt action category queries with the video-specific discriminated features, i.e., \(f_{h}(Q_1,\hat{X},\hat{X})\). The final output of query learner module is the learned action category query matrix \(\hat{Q}\) for the input video \(\textbf{x}\).\\

\noindent
\textbf{Query similarity Loss.}		To improve the learned action category queries and achieve better performance, we exploit the correlation between videos and propose a query similarity loss: For the \(k\)-th action category, we define a set \(V_k \) that contains all the videos in the training set that has this action in their ground-truth labels. For any two videos \(\textbf{x}_i\) and  \(\textbf{x}_j\) in \(V_k \), their learned action category query matrices are \(\hat{Q}_i\) and \(\hat{Q}_j\),  and the rows of these matrices \(\{\hat{\textbf{q}}^c_i\}_{c=1}^{C+1}\) and \(\{\hat{\textbf{q}}^c_j\}_{c=1}^{C+1}\) are the learned query vectors for C+1 categories, respectively.  Ideally, we would like the \(k\)-th category query vectors from these two sets, i.e., \(\hat{\textbf{q}}_i^k\)  and \(\hat{\textbf{q}}_j^k\),  to have similar representations, because they should contain the same abstract knowledge features for the \(k\)-th action category. The query similarity loss is defined as:
\begin{equation}
\label{eq:QS_loss}
\mathcal{L}_{QS} = \frac{1}{C+1}\sum_{k=1}^{C+1}\frac{1}{\binom{|{V_k}|}{2}}\sum_{\substack{\textbf{x}_i,  \textbf{x}_j \in V_k \\ \textbf{x}_i\neq  \textbf{x}_j}}d(\hat{\textbf{q}}_i^k, \hat{\textbf{q}}_j^k),
\end{equation}
\vspace{0.5cm}
where \(d(\textbf{e}_1, \textbf{e}_2)\) is the cosine distance:
\begin{equation}
\label{eq:cos_dist}
d(\textbf{e}_1, \textbf{e}_2) = 1 - \frac{\textbf{e}_1\cdot\textbf{e}_2}{\|{\textbf{e}_1}\|\|\textbf{e}_2\|},
\end{equation}
where \(\textbf{e}_1\) and \(\textbf{e}_2\) are two input vectors,  (\(\cdot\)) is the inner product and \(\|\cdot\|\) is the magnitude. The smaller the cosine distance is, the more similar the feature vectors are. 

\subsection{Query-Key Attention}
\label{sec:query-key-att}
Finally, for the input video \(\textbf{x}\), we have its final learned action category query matrix \(\hat{Q}\) and its video features \(\hat{K}\) (used as the final video key). As shown in \cref{fig:q-k}(b), each learned action category vector (a row of \(\hat{Q}\)) will be used to query on the video key \(\hat{K}\) at each time step by the vector inner product, and the output value is the attention weight of the corresponding action occurring at a time step. The higher the weight, the more likely that action occurs.
Our query-key attention operation is defined as:

\begin{equation}
\label{eq:q-k}
\psi(Q, K) = \frac{QK^\top}{\sqrt{D}},
\end{equation}
where \(Q\) and \(K\) are two input matrices. 

The temporal class activation map (T-CAM) \(A\) will be computed as:
\begin{equation}
\label{eq:T-CAM}
A = \psi(\hat{Q}, \hat{K}),
\end{equation}
which contains the attention weight for each action along the temporal dimension (\(T\)).  The softmax operation will be performed on T-CAM to calculate some training losses that we illustrate in  \cref{sec:train_objects}, \textit{e.g.}, the video-level classification loss. The effect of extremely small gradient will possibly be made after the softmax function, since the inner products could grow large in magnitude with a large value of \(D\). Therefore, as defined in \cref{eq:q-k}, we scale the value by \(1/\sqrt{D}\) to counteract this effect. 

\subsection{Training Objectives}
\label{sec:train_objects}
We adopt the top-k multiple instance learning strategy \cite{paul2018w} to compute the video-level classification loss. Given a training video \(\textbf{x}\), since we only have its video-level class ground-truth label, we will use the segment-level scores from its learned T-CAM \(A\) to first obtain the video-level class scores by aggregating the top k values along the temporal dimension for each class in \(A\), i.e., aggregating top k values in each row of \(A\):
\begin{figure}[t]
\centering
\includegraphics[width=1.0\linewidth]{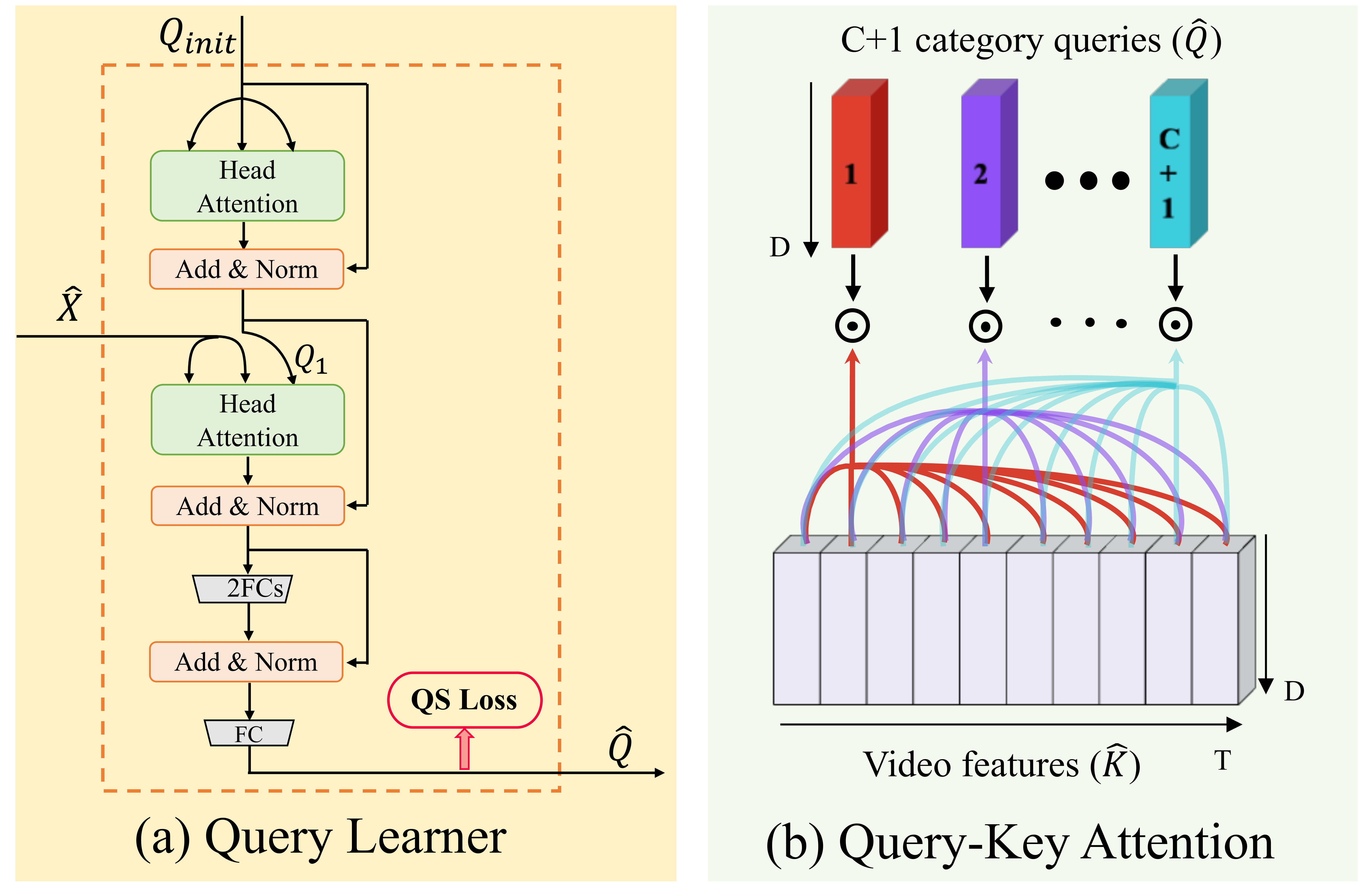}
\caption{(a) Query learner module. (b) Query-key attention module}
\label{fig4}
\label{fig:q-k}
\end{figure}

\begin{equation}
\label{eq:topk_mil_loss}
v_c = \frac{1}{k}\max_{\substack{{U\subset A_c} \\ |U| = k }}\sum_{i=1}^kU_i,
\end{equation}
where \(A_c\) is a set containing \(T\) attention weight values from the \(c\)-th row of \(A\). \(U_i\) is the \(i\)-th element in the set \(U\). We set \(k = \max(1, \lfloor{\frac{T}{m}}\rfloor)\), and m is a hyper-parameter. 

After that, we calculate the probability mass function (pmf) over all the action classes by applying softmax operation along the class dimension:

\begin{equation}
p_c = \frac{exp(v_c)}{\sum_{c^{'}=1}^{C+1}exp(v_{c^{'}})},
\end{equation}
where \(c = 1,2,...,C+1\).

The video-level classification loss is computed as the cross-entropy loss between the ground-truth pmf and the predicted pmf:
\begin{equation}
\label{VCLS_loss}
\mathcal{L}_{VCLS}^A = -\sum_{c=1}^{C+1}y_clog(p_c),
\end{equation}
where \([y_1,y_2,....y_C, y_{C+1}]\) is the normalized ground-truth vector, and the background activity is fixed to be a positive class since it always exists in the untrimmed videos. 

Following the previous work\cite{islam2021hybrid}, in order to better recognize the background activity and reduce its impact during inference, we apply the learned \(\textbf{s}\) (defined in  \cref{sec:main_pipe}) to suppress the background segments on the T-CAM \(A\) and obtain the background-suppressed T-CAM: \(\hat{A} = \) \(\textbf{s}\otimes A\), in which \(\textbf{s}\) element-wise multiplies on every row of \(A\). We then also calculate the video-level classification loss \(\mathcal{L}_{VCLS}^{\hat{A}}\) on \(\hat{A}\), and the background is fixed as a negative class now since it is suppressed. Our final video-level classification loss is denoted as: \(\mathcal{L}_{VCLS} = \mathcal{L}_{VCLS}^A  + \mathcal{L}_{VCLS}^{\hat{A}}\).

\begin{table*}[t]
\caption{The comparison with state-of-art TAL works on the THUMOS14 dataset. \dag refers to using additional information, such as human pose or action frequency. I3D is abbreviation for I3D features.}
\centering
\resizebox{2.00\columnwidth}{!}{
\begin{tabular}{c|c||ccccccc|ccc}
\toprule
\multirow{2}{*}{Supervision}  & \multirow{2}{*}{Method} & \multicolumn{7}{c|}{mAP@IoU(\%)}                          & \multicolumn{3}{c}{AVG mAP(\%)}  \\ 
\cline{3-12} 
\rule[0pt]{0pt}{10pt}                           & & 0.1  & 0.2  & 0.3  & 0.4  & 0.5  & 0.6  & 0.7 & 0.1:0.5 & 0.3:0.7 & 0.1:0.7      \\
\specialrule{0em}{2pt}{0pt}
\hline
\hline
\specialrule{0em}{2pt}{0pt}
\multirow{4}{*}{Fully}       
                             & TAL-Net \cite{chao2018rethinking} (CVPR'18)                & 59.8 & 57.1 & 53.2 & 48.5 & 42.8 & 33.8 & 20.8 & 52.3      & 39.8     & 45.1     \\
                             & GTAN \cite{long2019gaussian} (CVPR'19)                    & 69.1 & 63.7 & 57.8 & 47.2 & 38.8 & -    & -    & 55.3      & -        & -        \\
                             & VSGN \cite{zhao2021video} (ICCV'21)                    & -    & -    & 66.7 & 60.4 & 52.4 & 41.0 & 30.4 & -         & 50.2     & -        \\
                             & RefactorNet \cite{xia2022learning} (CVPR'22)                    & -    & -    & 70.7 & 65.4 & 58.6 & 47.0 & 32.1 & -         & 54.8     & -        \\
\specialrule{0em}{2pt}{0pt}
\hline
\specialrule{0em}{2pt}{0pt}
\rule[0pt]{0pt}{10pt}
\multirow{4}{*}{Weakly\dag}  & 3C-Net  \cite{narayan20193c} (ICCV'19)                 & 59.1 & 53.5 & 44.2 & 34.1 & 26.6 & -    & 8.1  & 43.5      & -        & -        \\
                             & PreTrimNet \cite{zhang2020multi} (AAAI'20)             & 57.5 & 54.7 & 41.4 & 32.1 & 23.1 & 14.2 & 7.7  & 41.0      & 23.7     & 23.7     \\
                             & SF-Net \cite{ma2020sf} (ECCV'20)                 & 71.0 & 63.4 & 53.2 & 40.7 & 29.3 & 18.4 & 9.6  & 51.5      & 30.2     & 40.8     \\
                             & BackTAL \cite{yang2021background} (TPAMI'22)                 & - & - & 54.4 & 45.5 & 36.3 & 26.2 & 14.8  & -      & 35.4     & -    \\
\specialrule{0em}{2pt}{0pt}
\hline
\specialrule{0em}{2pt}{0pt}
\rule[0pt]{0pt}{10pt}
\multirow{18}{*}{\makecell{Weakly \\ (I3D)}}     
                             & STPN \cite{nguyen2018weakly} (CVPR'18)            & 52.0    & 44.7    & 35.5 & 25.8 & 16.9 & 9.9 & 4.3  & 35.0         & 18.5     & 27.0        \\
                             & Nguyen \textit{et at} \cite{nguyen2019weakly} (ICCV'19) & 64.2    & 59.5    & 49.1 & 38.4 & 27.5 & 17.3 & 8.6  & 47.7         & 28.2     & 37.8        \\
                             & ACSNet \cite{liu2021acsnet} (AAAI'21)                & -    & -    & -    & 42.7 & 32.4 & 22.0 & -    & -         & -        & -        \\
                             & HAM-Net \cite{islam2021hybrid} (AAAI'21)                & 65.9 & 59.6 & 52.2 & 43.1 & 32.6 & 21.9 & 12.5 & 50.7      & 32.5     & 39.8     \\
                             & UM \cite{lee2021weakly} (AAAI'21)               & 67.5 & 61.2 & 52.3 & 43.4 & 33.7 & 22.9 & 12.1 & 51.6      & 32.9     & 41.9     \\
                             & FAC-Net \cite{huang2021foreground} (ICCV'21)                & 67.6 & 62.1 & 52.6 & 44.3 & 33.4 & 22.5 & 12.7 & 52.0      & 33.1     & 42.2     \\
                             & AUMN \cite{luo2021action} (CVPR'21)                & 66.2 & 61.9 & 54.9 & 44.4 & 33.3 & 20.5 & 9.0  & 52.1      & 32.4     & 41.5     \\
                             & CO$_2$-Net \cite{hong2021cross} (MM'21)            & 70.1 & 63.6 & 54.5 & 45.7 & 38.3 & 26.4 & 13.4 & 54.4      & 35.7     & 44.6     \\
                             & BaM+ACGNet \cite{yang2022acgnet} (AAAI'22)            & 68.1 & 62.6 & 53.1 & 44.6 & 34.7 & 22.6 & 12.0 & 52.6      & 33.4     & 42.5     \\
                             & MMSD \cite{huang2022multi} (TIP'22)                   & 69.7 & 64.3 & 54.6 & 45.0 & 36.4 & 23.0 & 12.3 & 54.0      & 34.3     & 43.6     \\
                             & DCC \cite{li2022exploring} (CVPR'22)                    & 69.0 & 63.8 & 55.9 & 45.9 & 35.7 & 24.3 & 13.7 & 54.1      & 35.1     & 44.0     \\
                             & FTCL \cite{gao2022fine} (CVPR'22)                   & 69.6 & 63.4 & 55.2 & 45.2 & 35.6 & 23.7 & 12.2 & 53.8      & 34.4     & 43.6     \\
                             & ASM-LOC \cite{he2022asm} (CVPR'22)               & 71.2 & 65.5 & 57.1 & 46.8 & 36.6 & 25.2 & 13.4 & 55.4      & 35.8     & 45.1     \\
                             & Huang \textit{et at} \cite{huang2022weakly} (CVPR'22)   & 71.3 & 65.3 & 55.8 & 47.5 & 38.2 & 25.4 & 12.5 & 55.6      & 35.9     & 45.1     \\
                             & DELU \cite{chen2022dual} (ECCV'22)   & 71.5 & 66.2 & 56.5 & 47.7 & \textbf{40.5} & 27.2 & 15.3 & 56.5      & 37.4     & 46.4     \\
                             & F3-Net \cite{moniruzzaman2023feature} (TMM'23)   & 69.4 & 63.6 & 54.2 & 46.0 & 36.5 & - & - & 53.9      & -     & -     \\
                             & ASCN \cite{moniruzzaman2023feature} (TMM'23)   & 71.4 & 65.9 & 57.0 & 48.2 & 39.8 & 26.8 & 14.4 & 56.4      & 37.2     & 46.2     \\
\cline{2-12}
\rule[0pt]{0pt}{10pt}
                             & \textbf{VQK-Net (ours)}    & \textbf{72.0} & \textbf{66.5} & \textbf{57.6} & \textbf{48.8} & 40.3 & \textbf{28.1} & \textbf{15.7} & \textbf{57.0}      & \textbf{38.1}     & \textbf{47.0}     \\
\bottomrule
\end{tabular}}
\label{tab:thumos14}
\end{table*}

As described in \cref{sec:main_pipe}, we adopt the mutual learning scheme \cite{hong2021cross} to learn the segment-level probabilities of being foreground action from both the RGB and flow input streams, and \(\textbf{s}_{rgb}\) and \(\textbf{s}_f\) should align with each other as they both represent the foreground probability of each segment along the temporal dimension \(T\), so a mutual learning loss is used as:
\begin{equation}
\label{eq:ml_loss}
\mathcal{L}_{ML} = \frac{1}{2}(\|\textbf{s}_{rgb} - \eta(\textbf{s}_f)\|_2^2  + \|\eta(\textbf{s}_{rgb}) - \textbf{s}_f\|_2^2),
\end{equation}where \(\|\cdot\|_2\) is the L2 norm, and \(\eta(\cdot)\)  stops the gradient of its input, so that \(\textbf{s}_{rgb}\) and \(\textbf{s}_f\) can be treated as pseudo-labels of each other. 

Based on the assumption that an action is detected from a sparse subset of the video segments \cite{nguyen2018weakly}, a sparsity loss \(\mathcal{L}_{Sparse}\) is used for the segment-level probabilities \(\textbf{s}_{rgb}, \textbf{s}_f\), and \(\textbf{s}\):
\begin{equation}
\label{eq:sparse_loss}
\mathcal{L}_{SP} = \frac{1}{3}(\|\textbf{s}_{rgb}\|_1 + \|\textbf{s}_f\|_1 + \|\textbf{s}\|_1).
\end{equation}

Moreover, since \(\textbf{s}_{rgb}, \textbf{s}_f, \textbf{s}\)  are the learned segment-level probabilities of being foreground action, they should oppositely align with the probability distribution of the background class, which is learned from the query-key attention operation, i.e., the \((C+1)\)-th row of \(A\) after it is applied by softmax operation along the column (class) dimension, denoted as \(\textbf{a} = column\_softmax(A)[C+1,\ :\ ]\in\mathbb{R}^T\). We use the guide loss \cite{islam2021hybrid} to fulfill this goal:
\begin{equation}
\label{eq:guide_loss}
\mathcal{L}_G = \frac{1}{3}(\|\textbf{1}-\textbf{a}-\textbf{s}_{rgb}\|_1+\|\textbf{1}-\textbf{a}-\textbf{s}_f\|_1 + \|\textbf{1}-\textbf{a}-\textbf{s}\|_1 ),
\end{equation}
where \(\|\cdot\|_1\) is the $l1$ norm, and \(\textbf{1}\in\mathbb{R}^T\) is a vector with all element values equal to 1.

We also adopt the co-activity similarity loss \(\mathcal{L}_{CAS}\) \cite{paul2018w} that uses the video features \(\hat{X}\) and suppressed T-CAM \(\hat{A}\) to better learn the video features and T-CAM \footnote{More details of the co-activity similarity loss a can be found in \cite{paul2018w}.}.

Finally, we train our proposed VQK-Net model using the following joint loss function:
\begin{equation}
\label{eq:final_loss}
\mathcal{L} = \mathcal{L}_{VCLS} + \alpha \mathcal{L}_{QS} + \mathcal{L}_{ML} + \beta \mathcal{L}_{G} + \mathcal{L}_{CAS}  + \gamma \mathcal{L}_{SP},
\end{equation}where \(\alpha, \beta\), and \(\gamma\) are the hyper-parameters.
\subsection{Temporal Action Localization}
\label{sec:temp_act_loc}
During testing time, given a video \(\textbf{x}\), we first calculate the video-level possibility of each action category occurring in the video from background-suppressed T-CAM \(\hat{A}\). We set a threshold to discard the categories with probabilities less than the threshold (set to 0.2 in our experiments). For the remaining action classes, we threshold on the segment-level foreground probability distribution \(\textbf{s}\) to get rid of the background segments and obtain the category-agnostic action proposals by selecting the continuous components from the remaining segments. We calculate the proposal's classification score \(\varepsilon\) by using the outer-inner score \cite{shou2018autoloc} on \(\hat{A}\). To enrich the proposal pool with proposals in different scale levels, we use multiple thresholds to threshold on \(\textbf{s}\). The soft non-maximum suppression is performed for overlapped proposals.

\section{Experiments}
\label{sec:experiments}

\subsection{Experimental Settings}
\noindent
\textbf{Datasets \& Evaluation metrics.}    We evaluate our approach on three widely used action localization datasets: THUMOS14  \cite{THUMOS14}, ActivityNet1.2 \cite{caba2015activitynet}, and ActivityNet1.3 \cite{caba2015activitynet}.

\textbf{THUMOS14} contains 200 validation videos and 213 test videos of 20 action categories. It is a challenging benchmark. The videos inside have various lengths, and the actions frequently occur (on average, 15.5 activity instances per video). We use the validation videos for training and the test videos for testing. 

\textbf{ActivityNet1.2} dataset has 4819 training videos, 2383 validation videos and 2489 test videos of 100 action classes. It contains around 1.5 activity instances per video. Since the ground-truth annotations for the test videos are withheld for the challenge, we utilize the validation videos for testing. 

\textbf{ActivityNet1.3} dataset has 10024 training videos, 4926 validation videos, and 5044 test videos of 200 action classes. It contains around 1.6 activity instances per video. Since the ground-truth annotations for the test videos are withheld for the challenge, we utilize the validation videos for testing. 

We evaluate our method with the mean average precision (mAP) at various intersections over union (IoU) thresholds. We utilize the officially released valuation code to calculate the evaluation metrics \cite{caba2015activitynet}.\\
\begin{table}[t]
\caption{Comparisons with state-of-art works on ActivityNet1.2 dataset. AVG means the average mAP from IoU 0.5 to 0.95 with step size 0.05.}
\centering
\resizebox{1.00\columnwidth}{!}{
\begin{tabular}{c|c||ccc|c}
\toprule
\multirow{2}{*}{Supervision} & \multirow{2}{*}{Method} & \multicolumn{4}{c}{mAP@IoU (\%)} \\
\cline{3-6} 
\rule[0pt]{0pt}{10pt}    &                         & 0.5    & 0.75   & 0.95  & AVG   \\
\specialrule{0em}{2pt}{0pt}
\hline
\hline
\specialrule{0em}{2pt}{0pt}
Fully                        & SSN \cite{zhao2017temporal}                    & 41.3   & 27.0   & 6.1   & 26.6  \\
\specialrule{0em}{2pt}{0pt}
\hline
\specialrule{0em}{2pt}{0pt}
\multirow{3}{*}{Weakly\dag}  
                             & SF-Net \cite{ma2020sf}     & 37.8   & 24.6   & 10.3   & 22.8  \\
                             & Lee \textit{et al} \cite{lee2021learning}     & 44.0   & 26.0   & 5.9   & 26.8  \\
                             & BackTAL\cite{yang2021background}     & 41.5   & 27.3   & 14.4   & 27.0  \\
\specialrule{0em}{2pt}{0pt}
\hline
\specialrule{0em}{2pt}{0pt}
\multirow{11}{*}{Weakly(I3D)}     
                             & DGAM \cite{shi2020weakly}                   & 41.0   & 23.5   & 5.3   & 24.4  \\
                             & HAM-Net \cite{islam2021hybrid}                & 41.0   & 24.8   & 5.3   & 25.1  \\
                             & UM \cite{lee2021weakly}                      & 41.2   & 25.6   & 6.0   & 25.9  \\
                             & ACSNet \cite{liu2021acsnet}                 & 41.1   & 26.1   & \textbf{6.8}   & 26.0  \\
                             & CO$_2$-Net \cite{hong2021cross}             & 43.3   & 26.3   & 5.2   & 26.4  \\
                             & AUMN \cite{luo2021action}                    & 42.0   & 25.0   & 5.6   & 25.5  \\
                             & BaM+ACGNet \cite{yang2022acgnet}             & 41.8   & 26.0   & 5.9   & 26.1  \\
                             & D2Net \cite{narayan2021d2}             & 42.3   & 25.5   & 5.8   & 26.0  \\
                             & CoLA \cite{zhang2021cola}             & 42.7   & 25.7   & 5.8   & 26.1  \\
\cline{2-6} 
\rule[0pt]{0pt}{10pt}        & \textbf{VQK-Net (ours)}   & \textbf{44.5}   & \textbf{26.6}   & 5.1   & \textbf{26.8}  \\
\bottomrule
\end{tabular}}
\label{tab:actnet1.2}
\end{table}

\begin{table}[t]
\caption{Comparisons with state-of-art works on ActivityNet1.3 dataset. AVG means the average mAP from IoU 0.5 to 0.95 with step size 0.05.}
\centering
\resizebox{1.00\columnwidth}{!}{
\begin{tabular}{c|c||ccc|c}
\toprule
\multirow{2}{*}{Supervision} & \multirow{2}{*}{Method} & \multicolumn{4}{c}{mAP@IoU (\%)} \\
\cline{3-6} 
\rule[0pt]{0pt}{10pt}    &                         & 0.5    & 0.75   & 0.95  & AVG   \\
\specialrule{0em}{2pt}{0pt}
\hline
\hline
\specialrule{0em}{2pt}{0pt}
\multirow{2}{*}{Fully}
                        & SSN \cite{zhao2017temporal}                    & 39.1   & 23.5   & 5.5   & 24.0  \\
                        & PCG-TAL \cite{su2020pcg}                    & 44.3   & 29.9   & 5.5   & 28.9  \\
\specialrule{0em}{2pt}{0pt}
\hline
\specialrule{0em}{2pt}{0pt}
\multirow{12}{*}{Weakly(I3D)}     
                             & STPN \cite{nguyen2018weakly}                   & 29.4   & 16.9   & 2.6   & -  \\
                             & TSCN \cite{zhai2020two}                & 35.3   & 21.4   & 5.3   & 21.7  \\
                             & UM \cite{lee2021weakly}                      & 41.2   & 25.6   & 6.0   & 25.9  \\
                             & ACSNet \cite{liu2021acsnet}                 & 36.3   & 24.2   & 5.8   & 23.9  \\
                             & AUMN \cite{luo2021action}                    & 38.3   & 23.5   & 5.2   & 23.5  \\
                             & TS-PCA \cite{liu2021blessings}                    & 37.4   & 23.5   & 5.9   & 23.7  \\
                             & UGCT \cite{yang2021uncertainty}                    & 39.1   & 22.4   & 5.8   & 23.8  \\
                             & FACNet \cite{huang2021foreground}                    & 37.6   & 24.2   & 6.0   & 24.0  \\
                             & FTCL \cite{gao2022fine}                    & 40.0   & 24.3   & \textbf{6.4}   & 24.8  \\
                             & Huang \textit{et at} \cite{huang2022weakly} & 40.6   & 24.6   & 5.9   & 25.0  \\
                             & MMSD \cite{huang2022multi}                    & 42.0   &  25.1   & 6.0   & 25.8  \\
\cline{2-6} 
\rule[0pt]{0pt}{10pt}        & \textbf{VQK-Net (ours)}   & \textbf{42.4}   & \textbf{26.4}   & 5.5   & \textbf{26.3}  \\
\bottomrule
\end{tabular}}
\label{tab:actnet1.3}
\end{table}
\noindent
\textbf{Implementation details.}		In this work, we sample the video streams into non-overlapping 16 frames segments and apply the I3D network \cite{carreira2017quo} pre-trained on Kinetics\cite{kay2017kinetics} to extract the 1024-dimensional segment-level RGB and flow features. For a fair comparison, we do not finetune the feature extractor. During the training stage, we randomly sample \(T=500\) segments for the THUMOS14 dataset and \(T=60\) segments for the ActivityNet1.2 and ActivityNet1.3 datasets. During the evaluation stage, all segments are taken. The values of  $\alpha$, $\beta$, and \(\gamma\) used in \cref{eq:final_loss} were determined experimentally. We found their optimal values to be: \(\alpha=5, \beta=0.8\), and \(\gamma=0.8\) for the THUMOS14 dataset, and \(\alpha=10, \beta=0.8\), and \(\gamma=0.8\) for  ActivityNet1.2 and ActivityNet1.3 datasets. To determine \(k\) in \cref{eq:topk_mil_loss}, m is set to 7 for the THUMOS14 dataset, 4 for the ActivityNet1.2 dataset, and 6 for the ActivityNet1.3 dataset. 

At the training stage, we sample 10 videos as a batch. In each batch, there are at least three pairs of videos such that each pair has at least one action category in common. We use the Adam optimizer \cite{kingma2014adam} with a learning rate of 0.00005 and weight decay rate of 0.001 for THUMOS14, a learning rate of 0.00003 and weight decay rate of 0.0005 for ActivityNet1.2 and ActivityNet1.3. For action localization, we use multiple thresholds from 0.1 to 0.9 with a step of 0.08, and we perform soft non-maximum suppression with an IoU threshold of 0.7. All the experiments are performed on a single NVIDIA Quadro RTX 8000 GPU.

\subsection{Comparison with State-of-art Methods}
In \cref{tab:thumos14}, \cref{tab:actnet1.2}, and \cref{tab:actnet1.3}, we compare our method with the existing state-of-art weakly-supervised methods and some fully-supervised methods. For the THUMOS14 dataset. We show mAP scores at different IoU thresholds from 0.1 to 0.7 with a step size of 0.1. Our VQK-Net model outperforms recent weakly-supervised approaches and establishes new state-of-the-art results on most IoU metrics. Moreover, our model outperforms some fully-supervised TAL methods and even some recent methods using additional weak supervisions, such as human pose or action frequency. For the ActivityNet1.2 and ActivityNet1.3 datasets, our method also reach state-of-art performance and outperforms some recent fully-supervised methods and the recent methods with additional weak supervisions. These results indicate the effectiveness of our proposed method.



\begin{table*}[t]
\caption{Evaluation of uniform and video-specific query learning strategies on THUMOS14.}
\centering
\resizebox{1.4\columnwidth}{!}{
\begin{tabular}{c|ccccccc|cc}
\toprule
\multirow{2}{*}{Exp} & \multirow{2}{*}{0.1} & \multirow{2}{*}{0.2} & \multirow{2}{*}{0.3} & \multirow{2}{*}{0.4} & \multirow{2}{*}{0.5} & \multirow{2}{*}{0.6} & \multirow{2}{*}{0.7} & \multicolumn{2}{c}{AVG mAP (\%)} \\
\cline{9-10} 
\rule[0pt]{0pt}{10pt} & & & & & & &                                 & (0.1:0.5)         & (0.1:0.7)
\\
\hline
\rule[0pt]{0pt}{10pt}
Uniform             & 69.7     & 64.3   & 55.6      & 46.7  & 39.0 & 26.2 & 13.8  & 55.1   & 45.1          \\
\rule[0pt]{0pt}{8pt}
Video-specific            & \textbf{72.0}    & \textbf{66.5}   & \textbf{57.6} & \textbf{48.8}  & \textbf{40.3}  & \textbf{28.1} & \textbf{15.7}   & \textbf{57.0} & \textbf{47.0}           \\
\bottomrule
\end{tabular}}
\label{tab:qk_strategy}
\end{table*}

\begin{table}[t]
\caption{Analysis of distance function used in query similarity loss on THUMOS14.}
\centering
\resizebox{0.7\columnwidth}{!}{
\begin{tabular}{c|cc}
\toprule
\multirow{2}{*}{Exp} & \multicolumn{2}{c}{AVG mAP (\%)} \\
\cline{2-3} 
\rule[0pt]{0pt}{10pt} & (0.1:0.5)    & (0.1:0.7)
\\
\hline
\rule[0pt]{0pt}{10pt}
Cosine             & \textbf{57.0}           & \textbf{47.0}                  \\
Jensen-Shannon     & 55.9           & 45.7 \\
Euclidean          & 55.1           & 45.1  \\
Manhattan          & 54.8           & 44.8  \\
\bottomrule
\end{tabular}}
\label{tab:distance}
\end{table}

\subsection{Ablation Studies \& Qualitative Results}
\label{subsubsec:different-qk}
\noindent
\textbf{Analysis on query learning strategies.}		
In the process of designing the query-key (q-k) attention mechanism, we investigate different strategies to learn our action categories queries, which is a key component in this mechanism. The performance of uniform and video-specific strategies was evaluated on the THUMOS14 dataset in \cref{tab:qk_strategy}. In the table, we first show the results of using the uniform strategy. In this experiment, the model does not include the video features in learning and learns a set of uniform action category queries for all the videos, i.e., the learned queries are not video-specific. The model simply relies on the learnable initial query embeddings, i.e., we do not use the query learner module (\cref{fig:q-k}(a)) in \cref{fig:framework}. The query similarity loss is not applicable in this case because it relies on the correlation of videos. 

While with the video-specific strategy, the model learns the video-specific action category queries, as described in \cref{fig:framework} and \cref{sec:query_learner}. From the table, it can be observed that the video-specific query learning strategy outperforms the uniform strategy quantitatively. 

\begin{figure}[t]
\centering
\includegraphics[width=1.0\linewidth]{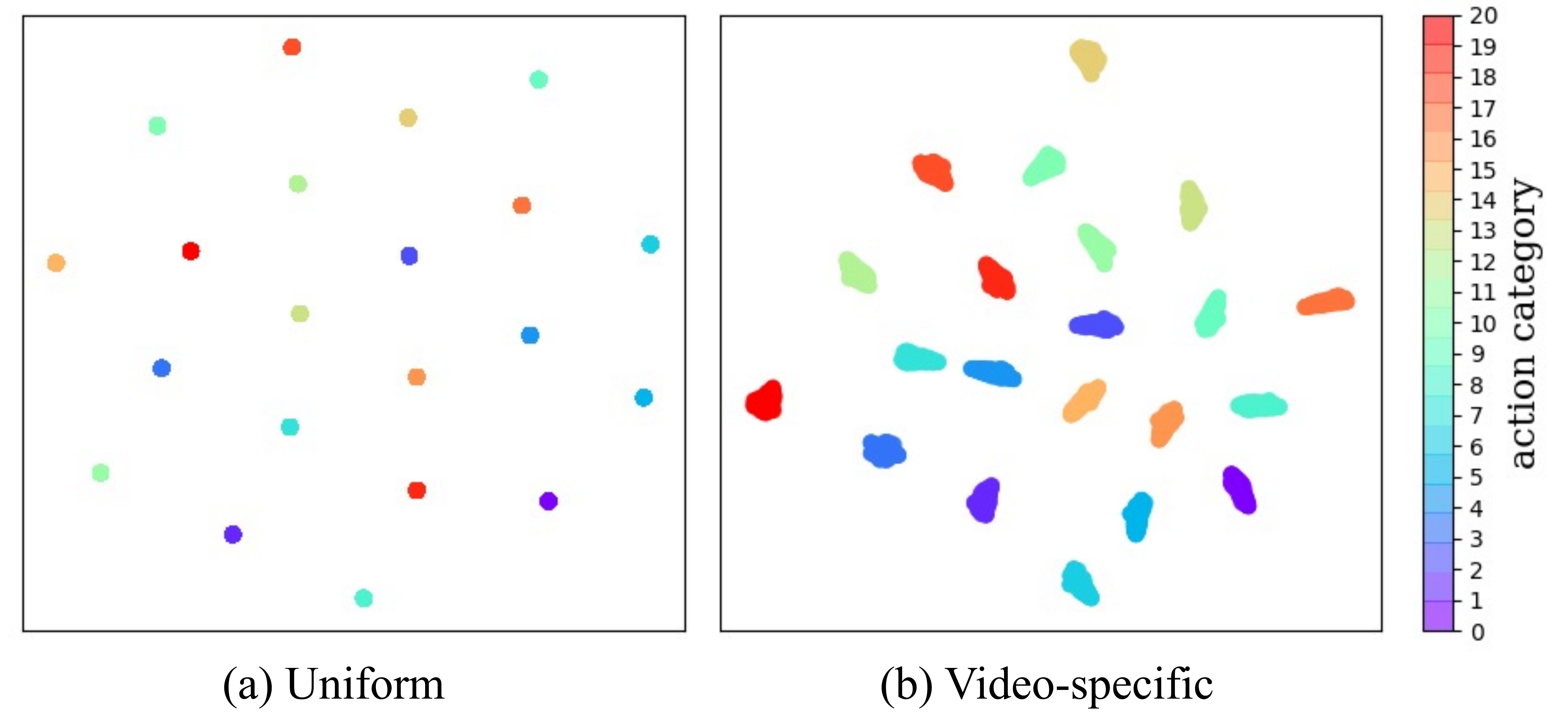}
\caption{Visualization of the learned action category queries on THUMOS14 test videos via t-SNE\cite{van2008visualizing}. }
\label{fig4}
\label{fig:tsne}
\end{figure}

\cref{fig:tsne}(b) shows the visualization of VQK-Net's learned action category queries for \(C\)+1 categories (including background) on test videos of THUMOS14 (\(C\) = 20), where the video-specific query learning strategy is used.  \cref{fig:tsne}(a) shows the learned action category queries using the uniform query learning strategy. From \cref{fig:tsne}(b), we can observe that there are 21 clusters of video-specific action queries for all test videos. This observation aligns with our hypothesis: the learned 21 category queries for each input video contain the abstract action knowledge features of 21 action categories, respectively, and compared to the uniform learning strategy where all videos have the same 21 action category queries (\cref{fig:tsne}(a)), video-specific action category queries have the ability to variant based on different input video scenes, to work optimally under the target video scenario while maintaining the core action knowledge features used to detect and identify actions in the target videos.
 
\begin{figure*}[t]
\centering
\subfloat[An example of CliffDiving action]{\includegraphics[width=2.0\columnwidth]{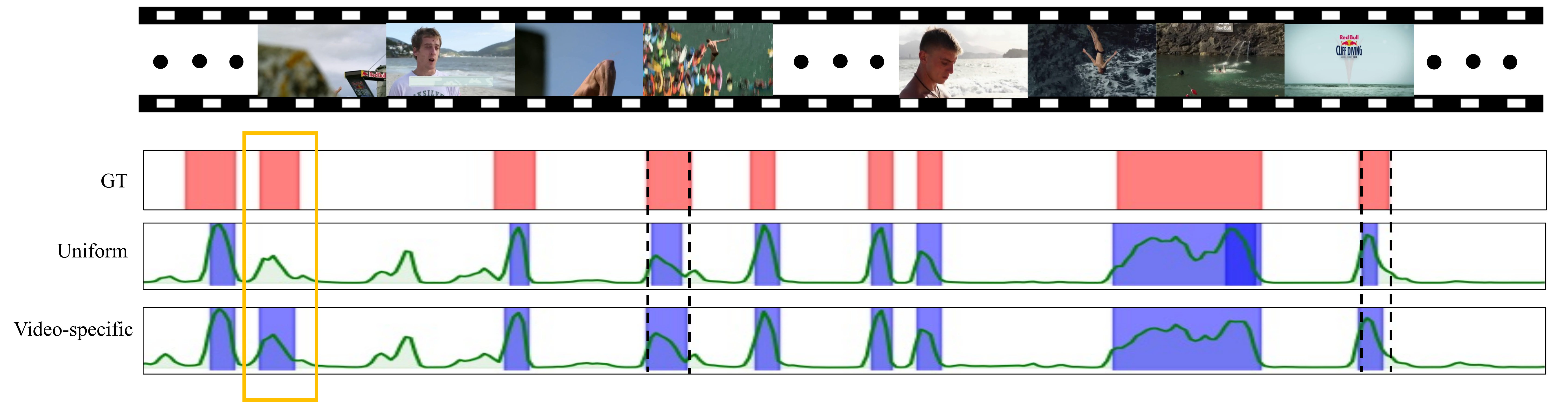}} \\
\subfloat[An example of HammerThrow action]{\includegraphics[width=2.0\columnwidth]{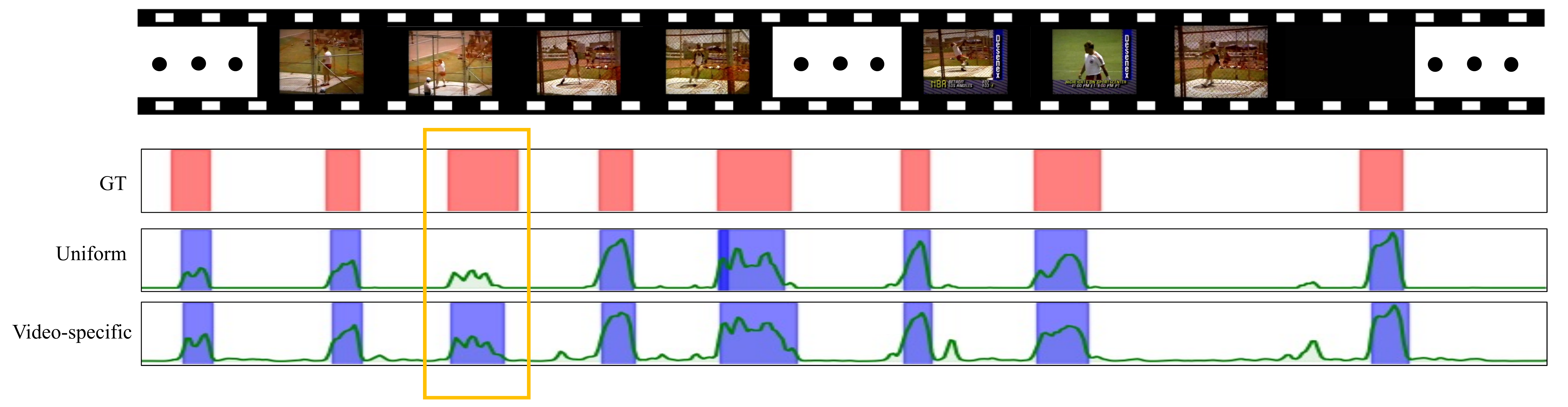}} \\
\subfloat[An example of ThrowDiscus action]{\includegraphics[width=2.0\columnwidth]{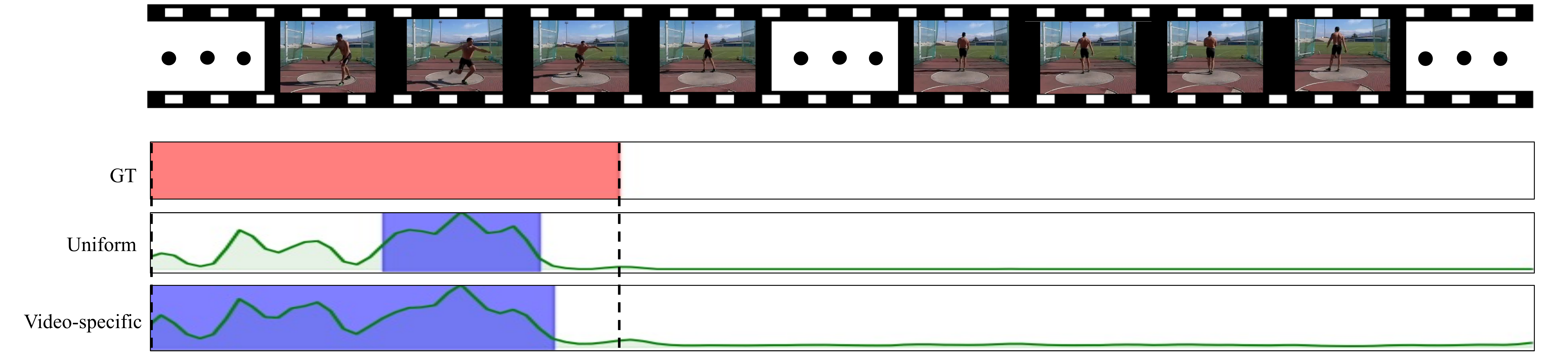}}
    \caption{Qualitative results on THUMOS14. The horizontal axis denotes time. The first plot is the ground truth (GT) action intervals. The remaining two plots illustrate the detection scores of ground truth action, shown in green curves, and the detected action instances using the uniform and video-specific query-key attention strategies, respectively.}
\label{fig:result}
\end{figure*}

We show some representative examples in \cref{fig:result}. For each example, the top row represents the ground truth localization. The uniform and video-specific correspond to the experiments from \cref{tab:qk_strategy}. From \cref{fig:result}, we can see that the video-specific query-key attention strategy predicts better localization against the uniform query-key attention strategy, demonstrating the effectiveness of the video-specific query-key attention modeling. Besides the increased precision in the localization, the video-specific approach can correct some missing detections from the uniform approach. In addition, even though some examples have frequent action occurrences, our VQK-Net model successfully detects all the action instances, which shows the ability to handle dense action occurrences.\\

\noindent
\textbf{Analysis of the distance function in query similarity loss.}	
In \cref{tab:distance}, we present the analysis of the distance function used in the query similarity loss (\cref{eq:QS_loss}). We can see that the cosine similarity distance performs the best, and the Jensen-Shannon distance is the second, while the Euclidean and Manhattan have a poor performance. This result aligns with the nature of our learned queries. Since the VQK-Net learns the video-specific action queries that could fit under different scenarios, the learned queries should not be precisely identical among different videos, as illustrated in the comparison in \cref{fig:tsne}. Therefore, using absolute distance such as Euclidean, Manhattan, etc., will not be appropriate.

\section{Conclusion}
\label{sec:conclusion}
In this paper, we propose a novel VQK-Net model that mimics how humans localize actions using video-specific query-key attention modeling. The VQK-Net learns video-specific action category queries that contain abstract-level action knowledge and can adapt to the target video scenario. We utilize these learned action categories to identify and localize the corresponding activities in different videos. We design a novel video-specific action category query learner worked with a query similarity loss, which guides the query learning process with the video correlations. Our approach shows the state-of-art performance on WTAL benchmarks.

{\small
\bibliographystyle{ieee_fullname}
\bibliography{egbib}
}
\end{document}